\documentclass{article}

\usepackage[preprint,nonatbib]{neurips_2022}

\usepackage[utf8]{inputenc} %
\usepackage[T1]{fontenc}    %
\usepackage[colorlinks]{hyperref}       %
\usepackage{url}            %
\usepackage{booktabs}       %
\usepackage{amsfonts}       %
\usepackage{nicefrac}       %
\usepackage{microtype}      %
\usepackage{xcolor}         %
\usepackage{amsmath}
\usepackage{amsthm}
\usepackage{graphicx}

\newcommand{\Dem}{D_\text{em}}

\title{AI Model Disgorgement: Methods and Choices}

\author{%
Alessandro Achille\enspace Michael Kearns\enspace Carson Klingenberg\enspace Stefano Soatto\\
AWS AI \\
{\tt \{aachille,kearmic,cklingen,soattos\}@amazon.com}
}

\date{February 26, 2023}

\begin{document}

\maketitle

\section{Introduction}

Responsible use of data is an indispensable part of any machine learning (ML) implementation. ML developers must carefully collect and curate their datasets, and document their provenance. They must also make sure to respect intellectual property rights, preserve individual privacy, and use data in an ethical way.

Over the past few years, ML models have significantly increased in size and complexity. These models require a very large amount of data and compute capacity to train, to the extent that any defects in the training corpus cannot be trivially remedied by retraining the model from scratch. Despite sophisticated controls on training data and a significant amount of effort dedicated to ensuring that training corpora are properly composed, the sheer volume of data required for the models makes it challenging to manually inspect each datum comprising a training corpus. 

One potential fix for training corpus data defects is model disgorgement --- the elimination of not just the improperly used data, but also the \emph{effects} of improperly used data on any component of an ML model. Model disgorgement techniques (“MD” hereafter) can be used to address a wide range of issues, such as reducing bias or toxicity, increasing fidelity, and ensuring responsible usage of intellectual property. Developers of large models should understand their options for implementing model disgorgement, and novel methods to perform model disgorgement should be developed to help ensure the responsible usage of data. At the same time, developers should help content owners and creators understand how their data affect trained models. Content owners can then incorporate such knowledge into licensing terms for the lawful usage of their data, or otherwise specify terms to protect their data from improper usage.

In this paper, we introduce a taxonomy of possible disgorgement methods that are applicable to modern ML systems. In particular, we investigate the meaning of “removing the effects” of data in the trained model in a way that does not require retraining from scratch.

One of the core challenges with MD for large models is that they contain parameters representing an arbitrarily high number of data dimensions and statistical correlations. This means that it is very difficult to determine the specific effect that any particular piece of training data had on a fully trained model. For this reason, and others discussed later, it is not a viable strategy to “just delete the data”. 

This paper presents both novel and known technical approaches to MD. These approaches are distinguished along three dimensions:
\begin{enumerate}
    \item \textbf{Disgorgement certificate type.} We must consider what it means to “remove the effects” of particular data on a model. Here, we draw a technical distinction between \textit{deterministic} processes (where we can say that particular data has categorically not affected a model) versus \textit{probabilistic} processes (where we can say that the effect of particular data on a model is essentially zero or \textit{de minimis}). 
    \item \textbf{Disgorgement verification type.} Here we make a distinction between methods for disgorgement that are structural and those that are \textit{empirical}. By structural we mean methods that alter the ML workflow structure in a way that enforces disgorgement, or facilitates later disgorgement. An example of the latter would be a model architecture that averages many smaller sub-models, each trained on different partitions of the data; disgorgement of any particular portion of data is then achieved by dropping its corresponding sub-model from the average. By empirical we mean methods that aim for approximate disgorgement, and for which verification might require empirical measurement, or numerical quantification of the degree of disgorgement. Such methods include various “forgetting” or “unlearning” approaches as well as differential privacy.
    \item \textbf{Disgorgement temporal application.} We also distinguish approaches based on when those techniques must be applied as part of the model composition process. We present approaches that can be applied reactively (applied {\em post hoc} to a fully trained model), proactively (where optionality must be proactively reserved as part of the training process), or preemptively to reduce or eliminate the need for disgorgement thorough a properly designed training process.
\end{enumerate}

The techniques we will discuss are listed below:

\begin{table}[h]
    \centering
    \begin{tabular}{ccc}
    \toprule
         & \textbf{Deterministic} & \textbf{Probabilistic} \\
         \midrule
        \textbf{Reactive} & Retraining & Forgetting/Unlearning \\
        \textbf{Proactive} & Compartmentalization & - \\
        \textbf{Preemptive} & Dataset Emulation & Differential Privacy\\
        \bottomrule
    \end{tabular}
\end{table}

\section{Reactive Disgorgement}

Large-scale neural networks can be trained on datasets that comprise web-scale data.  In such cases, the need for model disgorgement may arise from, among other things, errors in the data collection and curation process.

\paragraph{Retraining.} The naive method to remove the influence of any cohort of data from a trained ML model is to retrain the model from scratch using only the remaining data. However, considering that neural networks currently in use can have hundreds of billions or even trillions of parameters, retraining the model from scratch after each MD request is not a viable approach. Apart from cost considerations, when the ML model is used as part of a workflow in a larger system, the retrained model is generally not compatible with the original one \cite{shen2020towards,srivastava2020empirical}. Slight perturbations of the training process, even if retraining the same model on the same data, can yield models whose behavior is sufficiently different to disrupt the behavior of downstream workflows \cite{bansal2019updates,yan2021positive}. Thus, naive MD based on retraining would render most large-scale ML systems essentially unusable. 

\paragraph{Selective Forgetting/Unlearning.} If the cohort of data that triggers the need for MD is a small fraction of the overall training data, it may be possible to characterize and eliminate its influence on the trained model without the need to retrain the model. This approach is referred to in the literature as machine unlearning \cite{cao2015towards,ginart2019making} or selective forgetting \cite{Golatkar_2020_CVPR,mixedprivacyforgettinggolatkar,golatkar2020forgetting}. Due to the complexity of large-scale deep networks, estimating the influence of a sample in a way that is simultaneously efficient and precise is challenging \cite{achille2019information}. Hence, such procedures cannot provide deterministic removal guarantees, but can provide quantifiable probabilistic measures. We will discuss the meaning of such probabilistic measures later in this paper. Many approaches to unlearning are based on the notion of influence functions \cite{koh2017understanding}, which provide methods to efficiently estimate the influence of selected samples through an approximation of the loss function used to train the ML model. Such probabilistic measures, however, fail when applied to large-scale deep networks due to the complexity of the training process and the highly non-convex nature of the loss function \cite{basu2020influence}. Some of these challenges can be mitigated if the developer of the ML model has at its disposal a “safe core” set of training data that is known to never require disgorgement, for instance synthetic data generated \textit{ab ovo}. In particular, the safe core set can be used to initialize the model, and information from additional data can be encoded as a small perturbation of the core model. This allows to more easily estimate the influence of additional data, and to remove their influence if needed without catastrophic consequences to the overall ML model \cite{mixedprivacyforgettinggolatkar}. 

\section{Proactive Disgorgement}

Given that even the tightest standards of data curation can be imperfect at the scale of the datasets in use today, it may be worthwhile to train ML models in preparation for possible MD. That is, the ML models could be trained in such a way that MD can be later performed with minimal impact on the overall trained model. We call this proactive disgorgement.

A conceptually simple approach to proactive disgorgement \cite{sisa,arcane,legonet,kochno,kumar2022privacy} involves splitting the dataset into multiple disjoint subsets (shards) and training separate models in isolation on each shard. Then, disgorging a sample only concerns the sub-models trained on the subset containing the sample. Disgorgement requests can be addressed simply by eliminating or retraining the components of the model that have been exposed to the cohort of data in question.  (If these components are trained preemptively, as we explain below, even that step may be unnecessary.)

We refer to these methods as “compartmentalization”, as they separate information from different samples in different sub-models. The main design choices for compartmentalized models concern how to split the training data, what architecture to use for the sub-models and how to combine them. These in turn affect three aspects of the trained ML system: expected forgetting cost, model accuracy, and inference time latency.

Each time a forgetting request is received, the sub-model corresponding to the shard that is being disgorged needs to be retrained, and the retraining time is roughly proportional to the size of the shard. Hence, using smaller shards (especially for samples that have a higher-risk of being disgorged) leads to lower expected computational cost for each forgetting request. On the other hand, training sub-models on increasingly smaller shards results in progressively weaker models, thus reducing final accuracy. Moreover, splitting the data in small shards results in a larger total number of shards, each associated to a sub-model that needs to be stored and executed at inference time, increasing the inference latency.

\begin{figure}
    \centering
    \includegraphics[width=0.5\linewidth]{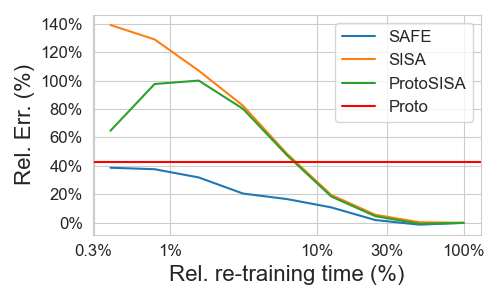}
    \caption{\textbf{Example of Pareto curves for different forgetting algorithms} (reproduced from \cite{bowman2023safe}). Models trained on 7 fine-grained visual classification tasks using different compartmentalization methods. We report the average (normalized) test error (y-axis) across tasks when changing the budgeted time allowed to forget a sample (x-axis, normalized with respect to total time to retrain from scratch). Lower curves are better. Compartmentalization allows significant reduction in the cost to satisfy a forgetting request (up to only 0.3\% of the baseline retraining from scratch). This, however, comes at the cost of increased error with respect to a paragon model trained without compartmentalization (rightmost point). Users can select the best trade-off for their applications, but better methods can improve the trade-off.}
    \label{fig:pareto}
\end{figure}

The user then needs to choose the desired trade-off between expected forgetting cost and accuracy (and latency) along a Pareto curve portrayed in Fig.~\ref{fig:pareto}. Better design of the model disgorgement can however lead to better Pareto curves.

SISA \cite{sisa} is a baseline compartmentalization method that randomly splits the data into uniform shards and trains a separate copy of a standard network on each. ARCANE \cite{arcane} suggests starting with a network pretrained on a core dataset deemed safe from disgorgement requirements, and constructs a simple linear classifier on the data by computing the average embedding. This leads to lower accuracy than an unconstrained model, but enables \textit{instant forgetting}: to remove a sample, it is sufficient to subtract its embedding from the mean, without any retraining (thus the expected forgetting cost is close to zero). ARCANE has better accuracy than SISA when low forgetting cost is required. SISA can also be applied to connectors learned on a pretrained backbone, as demonstrated in \cite{kumar2022privacy} on natural language processing with up to 14 shards. SAFE \cite{bowman2023safe} improves over SISA along all three axes: (a) it does not use uniform sharding, but rather optimizes the shard composition to improve the sub-models’ performance; (b) uses a similar mechanism to ARCANE to ensure good accuracy for extremely low retraining time, (c) uses InCA adapters \cite{inca} as sub-models, which allows training and running inference in parallel on thousands of shards with a significantly reduced computational cost. LegoNet \cite{legonet} focuses on improving the ensembling procedure, and selects and averages the top-$k$ most relevant sub-models in an instance-dependent manner.

A further advantage of compartmentalization-based approaches is that disgorgement of data of an entire shard (or multiple shards) --- as opposed to disgorgement of individual samples --- can be performed at essentially zero cost. Hence, if one expects that all samples from a data source may need to be disgorged at the same time, it may be beneficial to group all data from that source into a single shard, which may then easily be dropped. This, however, may be difficult to realize if the source does not exhibit sufficient variety to enable training strong sub-models. For example, if shards were organized by domain, each model built on a homogeneous shard would overfit to that domain, resulting in a collection of biased models. Combining such models would likely lead to a severe performance degradation compared to an unconstrained model trained on a monolithic dataset.

Compartmentalization ensures that there is strictly zero influence of the disgorged data, although at an uncertain cost of disgorgement since the data to be disgorged is not known ahead of time. Here, cost refers to not just the computational cost of retraining the affected sub-models, but also to the loss in accuracy due to architectural constraints and nature of the shards, which may in some cases be so significant as to render the ML model ineffective. In the limit where the data to be disgorged is distributed across all shards, this method requires retraining from scratch. In the next section, we discuss the alternative approach, which is to make the cost of disgorgement zero, by preempting the need to disgorge, in exchange for a probabilistic guarantee on the influence of the cohort of data in question.

\section{Preemptive Disgorgement}

Preemptive disgorgement refers to modifications of the training process that, by design, ensure that ``unique information'' \cite{harutyunyan2021estimating} contained in any cohort of samples in the training data is bounded by a small value selected by the system designer. Since no substantial information about any training sample is present in the model, in principle nothing needs to be done to satisfy an individual disgorgement request. However, as we will elaborate later, disgorgement of larger groups of data still present a challenge. In the next section, we describe the main elements of the framework of differential privacy (DP). In the following section, we discuss challenges that arise when applying DP to high dimensional data whose variability can almost entirely be ascribed to nuisance factors, such as images. Finally, we make recommendations to combine compartmentalization and DP to address the most challenging cases.

\subsection{Differential Privacy}

In many ways, \textit{differential privacy} (DP) can be considered the “gold standard” of model disgorgement, in that it proactively trains a model in a way that provides a mathematical proof that no particular piece of training data had more than a negligible effect on the model or the content it generates. The technical details of the definition of DP are beyond the scope of this document, but its force is captured in the following thought experiment. Imagine the original training dataset D, and consider the dataset $D'$ that results from removing any small fraction of $D$ (say, all the works of a particular artist). Then under DP, a model trained on $D$ is (provably) statistically indistinguishable from one trained on $D'$ \textit{even to an observer who knows both $D$ and D’}. In the generative setting, this means the distribution over output content for a given input prompt is also indistinguishable. In other words, a DP model effectively already disgorges any (small) amount of training data by minimizing its effects \textit{a priori}.

DP models are achieved by deliberately adding noise to the training process, in a way that attempts to eradicate the impact of any small piece of the data while still having the desired aggregate effects (in generative models, high quality outputs). In general one expects that adding more noise provides stronger disgorgement properties but will also degrade output quality. This trade-off is determined by a privacy parameter that must be tuned and chosen. 

The practice of DP model training remains in its infancy, and its effectiveness is untested on the scale of the hundreds of billions of parameters common in modern generative models. We should expect there to be challenges to the adoption of DP as a disgorgement solution, but it is a worthwhile standard for comparison for other approaches, and may eventually be a feasible solution. Moreover, DP can be combined with other methods, such as compartmentalization, and leverage a “safe set” to yield a more favorable privacy/performance trade-off \cite{golatkar2022mixed}.

\subsection{Dataset Emulation}

Some disgorgement requests arise in the context of “generative AI”, where the concern is that the ML model may generate data that is “similar to”, “in the style of”, or “captures the spirit of” data used for training. The goal of dataset emulation is to generate an “emulated” synthetic dataset  $\Dem$ that captures the general distributional properties of the  original training set $D$ while maximizing the \textit{geometric, perceptual, or conceptual distance} from $D$, defined below. 

\begin{figure}[h]
    \centering
    \includegraphics[width=0.9\linewidth]{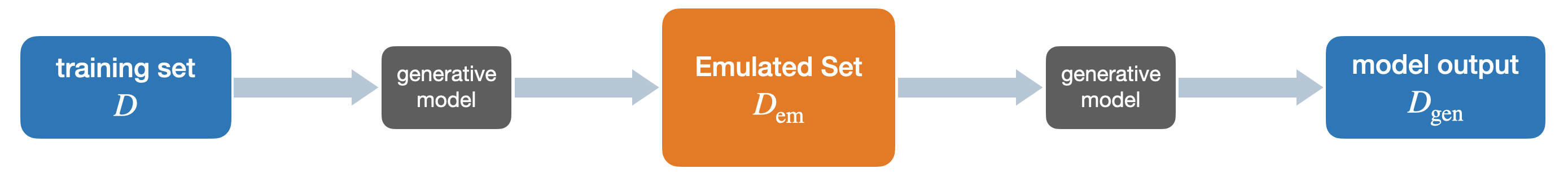}
    \label{fig:dataset_emulation_diagram}
\end{figure}

\paragraph{Clean Room.} While a dataset $\Dem$ generated from a model trained on $D$ has a computational relationship with the latter, that relationship can be limited to capture general distributional properties while exercising maximum care to steer clear of generating samples close to $D$. Once such a guarantee is provided, the data $\Dem$ should be usable in downstream tasks without implicating the protectable elements of $D$. In particular, the resulting data can be used to train a model, now without constraints, that can be used to discriminate or generate new outputs $D_\text{gen}$. In the latter case, the emulated dataset $\Dem$ acts as a “clean room” to separate the model that generates $D_\text{gen}$ from the dataset $D$, which not only has never been seen by the model that produces $D_\text{gen}$, but which has samples that are, by design, as different as possible from it. If so desired, users can manually inspect $\Dem$ to ensure that transferred elements are not contained in it, or that it is sufficiently different, before the final model is trained.

\paragraph{Generating $\Dem$.} There are essentially infinitely many models that could be trained on $D$, each involving design choices and randomness, and each able to generate essentially infinitely many different datasets $\Dem$.  While a model trained on $\Dem$ may contain significant novel elements, the owner of $D$ may still want to protect specific aspects implicit in $D$. These aspects can be measured directly on each sample of the emulated dataset $\Dem$ and removed automatically or manually so that the latter is guaranteed to not have such information. Alternatively, one can use a generative model trained on $D$ to create synthetic data $\Dem$ that structurally is biased to differ as much as possible from $D$ on these specific aspects. This bias can be hard-coded in the loss function used to train the model, rather than verified from the output of the model. This can be achieved by training the generative model while also maximizing some distance $d(x,x')$ between the generated samples $x'$ and the training samples $x$. Of course, the same distance can be used to reject generated samples {\em post hoc} as an additional safety measure, for instance if  $d(x,x')<\epsilon$ for a tunable $\epsilon$. We note, however, that training a generative model on entirely synthetic data remains an open challenge: even if the images in $\Dem$ are realistic and subjectively “similar” to those in $D$, the quality of an ML model trained on $\Dem$ is typically substantially degraded compared to one trained on $D$ (see below). This is because synthetically generated images, even when realistic, usually contain subtle artifacts that affect the training process. Such artifacts impact the quality of both generative and discriminative models \cite{zhao2020sim,xie2020unsupervised}, sometimes referred to as the “sim-to-real gap.”

Below we show an example of synthetically generated data, where the conceptual distance (defined below) is chosen to be the Euclidean distance among pre-trained CLIP embeddings. Below we show some of the generated synthetic samples in $\Dem$, as well as the closest samples in $D$. We see that the $\Dem$ is composed of substantially novel images and not just imitations of the original training data. We then use $\Dem$ to train a model on a downstream flower classification task. The model trained on $\Dem$ achieves an accuracy of 52\%, catastrophically lower than what would be achieved training directly on D (96\%) due to the above-mentioned sim-to-real gap, corresponding to a 1200\% error increase. At the present state of technology, therefore, an emulated dataset may be viable if it is the end product for human fruition, but is not yet viable as a substitute for the original dataset to train a generative or discriminative model. However, we expect steady progress in the area of  Unsupervised Domain Adaptation, so we consider this technique “potentially viable” given the speed of evolution of the field.

\begin{figure}[h]
    \centering
    \includegraphics[width=0.7\linewidth]{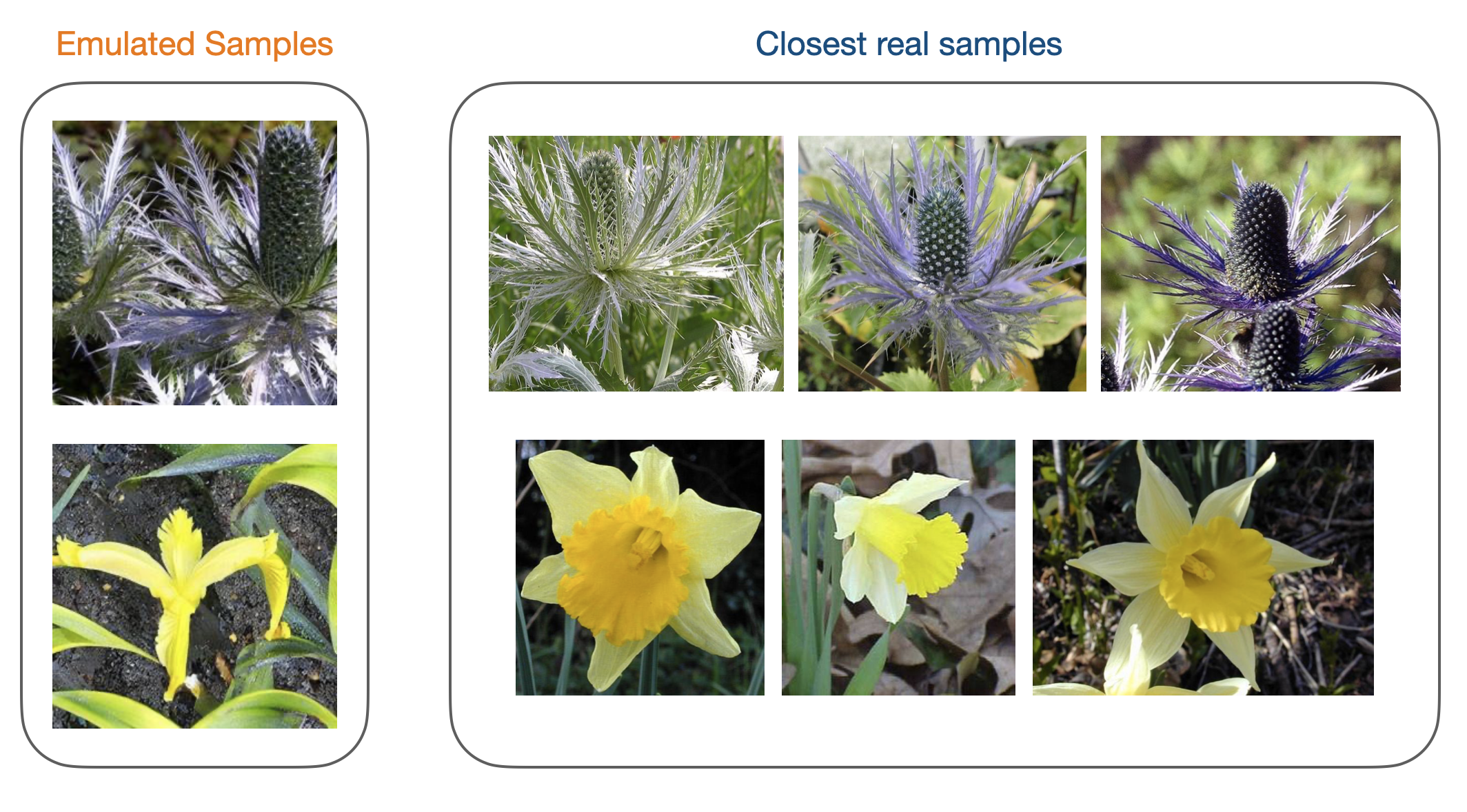}
    \label{fig:dataset_emulation_experiment}
\end{figure}

\textbf{Separation through descriptions.} Yet another option, is for $\Dem$ to be generated in a different modality than $D$. For example, $\Dem$ could consist of restricted abstract encodings (embedding vectors not directly renderable as images), or encodings mapped to semantically interpretable values, for instance captions or text embeddings. In this case, the value of the emulated dataset for downstream training is even further reduced.

\textbf{On the choice of distance.} For symbolic data such as text (represented either in raw form or as a vector embedding $x$), there are objective (``geometric'') distances $d_g(x_1, x_2) = \|x_2 - x_1\|_p$  that well approximate perceived similarity. However, for signal data such as images, any datum $x_1$ has a large number of samples that are far from $x_1$
according to $d_g$, yet are perceptually indistinguishable. This can occur both for subtle changes \cite{wu2020unsupervised} or for macroscopic ones \cite{rensink2001change}. Indeed, one can significantly change the value of every pixel in an image without triggering a perceptible change. To capture perceptual, rather than geometric, similarity, one can compare low-level statistics $\phi$ designed to mimic the early stages of human cortical processing, designed so that data points that have small distance are effectively (with high probability) indistinguishable by humans \cite{wang2004image}. In this case, we write the “perceptual distance” as  $d_p(x_1, x_2) = \|\phi(x_1) - \phi(x_2)\|$ which shifts the comparison from data space to feature space, where a standard geometric distance is applied. Beyond a pre-designed feature space that depends on the sensor but not the object being sensed, one consider the abstract concept of  “style” or “spirit” of a certain dataset. In this case, one would have to define not a geometric distance $d_g$ or perceptual distance $d_p$, but a “conceptual distance” $d_c(x_1, x_2) = \|\psi(x_1) - \psi(x_2)\|$ where now $\psi$ is no longer designed to capture low-level statistics, but rather it is trained to capture high-level semantic characteristics of the data. For instance, $\psi$ can be the embedding produced by a model trained to align images and their corresponding textual explanations, or ``captions''  \cite{radford2021learning,heusel2017gans}. 

\noindent{\bf Giving control to content owners and creators.} Note that a choice of distance defines what is to be considered tantamount to reproduction of the data beyond verbatim reproduction of the data itself (induction).  If the distance is lax, it may cover a portion of data space far beyond the original data, possibly including data belonging to other owners. For example, say a vendor who owns $D$ is interested in licensing it for training to a service provider who already has an ML model trained on the dataset $\Dem$. The vendor may require that all data generated be farther than some $r>0$ from $D$ according to an adversarially chosen distance $d$. Clearly, if the images already generated by the service falls within a distance $r$, despite having been obtained without any knowledge of $D$, then the owner of $D$ would be attempting to control the usage of data that does not belong to them. Thus, the validity of the exercise falls in the choice of the distance or discriminant that determines protectable elements of the data $D$. Such a distance or discriminant would have to be carefully tested and calibrated so as not result in over-reaching. Even so, a distance could be fooled into misclassifying data as being sufficiently distant by applying imperceptible perturbations \cite{moosavi2017universal}, although distances can be devised that are robust to such perturbations \cite{zeng2023towards,shan2023glaze}. 

In general, the “essence” or “style” of a dataset $D$ is an abstract concept not included in the set of ``protectable elements'' of $D$, with models trained on it representing different embodiments, each comprising novel elements that are not in the specification of the dataset. These novel elements define the inductive bias needed to generate novel samples from the training data $D$. There are infinitely many possible embodiments of ML models trained using $D$, each of which can have different inductive behavior, each of which can generate different novel data. It is not possible to test each embodiment, since there are infinitely many. Moreover, even if it was possible for an ML model trained on $D$ in finite time with finite memory to “capture” the abstract concept of the “spirit” or “essence” or “style” embodied in $D$, whether such capture has actually happened is not decidable with any finite empirical test \cite{achille2022binding}. 

For high dimensional data such as images, Dataset Emulation provides an alternative means to ensure that synthetic data is sufficiently different, which can be done constructively by designing the learning criterion in such a way that it maximally differentiates synthesized samples from the original ones, or selectively by testing each synthesized sample and rejecting it deterministically if it fails the differentiation test. Challenges remain in this approach, however, since synthetic data --- even if realistic and indistinguishable from real to the human eye --- contain subtle artifacts that affect performance of trained models. This is the analog of the “uncanny valley” of physics-based synthesis, but transposed to an inductive learning setting. One of the key elements of Dataset Emulation is the choice of the criterion (distance or discriminant) that inductively defines the “style” or “spirit” of a particular dataset. In our framework, this choice is delegated to the function $\psi$, left unspecified, since there is no canonical choice for it, and which choice is selected may depend on the specific application, dataset, or other factors. The choice of distance, as well as the margin, may be part of the dataset cards and factor into the conditions under which the data is used. The set of options should be at the disposal of content owners and creators when negotiating the terms of use for their data, thus giving them control of how their data is used, how it affects the trained model, and how its contribution can be duly compensated.

\section{Discussion}

The methods we have surveyed vary in their applicability and value, depending on the intended application and desired protection from the need to disgorge data. Below we discuss some important considerations:

\paragraph{Preemptive DP and Group Forgetting.} Consider the case where a group of related data points needs to be disgorged from a model. While DP ensures that the influence about any individual sample is negligible, the bound on the influence can quickly become vacuous as the size of the group increases. However, how this should be interpreted depends on the composition of the group. Depending on the size of the group and the initially selected privacy level, it may not be possible to guarantee negligible influence via DP as measured at the data group level. Notions such as user-level DP aim to address this problem, by grouping data of the same entity during training. However they require {\em a priori} knowledge during training of which samples are grouped, and what the composition of those groups are. For large-scale data usage, being able to group images along each possible grouping dimension may not be realistic, and no single grouping may be correct. This suggests using preemptive methods together with proactive methods (see next paragraph): the first minimizes the number of retrainings, the latter allows to retrain only smaller portions of the models.

\paragraph{Compartmentalization and Group Forgetting.} Compartmentalization methods are particularly well-suited to scenarios where a significant portion of data needs to be disgorged together. For example, the terms of a limited data license may require an entire dataset to be disgorged after a period of time, where each of the many images in the cohort has some shared information, for instance a watermark. If the entire cohort was compartmentalized in a single shard, then disgorgement is trivially accomplished by removing that shard and every sub-model that used it during the training process. Provided that the model was originally trained with a sufficiently large “safe core” set of data, the impact on performance should remain limited. The key issue in disgorgement is how to shard the data. There is a vast design space that affects the potential cost of disgorgement and impact on the model performance. In addition to performance, model bias is also a concern, especially if shards are segregated by class or domain, for the resulting models would overfit and simplistic ensembling of sub-model activations are not likely to be sufficient to correct such biases. For this reason, shards should be sufficiently diverse to ensure the quality of sub-models trained on them. We note however that, while better sharding may decrease the cost of forgetting and improve model performance, compartmentalization provides guaranteed deterministic forgetting regardless of the sharding used.

\paragraph{Interpretation of negligible influence.} Both probabilistic unlearning and preemptive DP rely on bounds on the remaining influence of samples. For example, $(\epsilon,\delta)$-DP bounds the influence of individual samples through two tunable parameters $\epsilon$ and $\delta$ (the discussion for probabilistic unlearning and forgetting is similar). Smaller $(\epsilon,\delta)$ correspond to less influence, but it may not always be straightforward to translate this into user-facing measures. One strong guarantee is that, if $D$ and $D'$ are two datasets differing by an example,  an observer who knows both cannot design a statistical test at significance level $\alpha$ with power greater than $e^\epsilon \alpha + \delta$ ({\em i.e.}, as mentioned above, one cannot determine with confidence if a sample was used to train the model or not) \cite{dong2022gaussian}. Another important user-facing metric is how similar images or text generated by a model trained with or without a sample would be. While it remains true that smaller $(\epsilon,\delta)$ always lead to more similar images, we also note that exact quantification of similarity is difficult since human conceptual perception of similarity (especially in images) does not relate directly to metric similarity (e.g., difference in pixel values or log-likelihood of the output). In particular, exceedingly small values of $\epsilon$ may be needed to ensure conceptual similarity.

\paragraph{Interpretation of forgetting and model outputs.} Yet another consideration is the mismatch between what it means to forget a sample and what the user may expect from the resulting model. A disgorged model may still produce outputs that are similar (or conceptually similar) to the disgorged data. This is not a bug: for example, a string of text may be the likely answer to a prompt regardless of whether it was observed during training. One could consider, however, being more proactive and avoiding outputs too similar to disgorged data (even if this would, paradoxically, make the model less private with respect to that data). Dataset Emulation follows a similar approach.

\paragraph{Preemptive methods, model performance, long tails and fairness.} By design, preemptive methods reduce influence of any examples. This has some unavoidable consequences on the performance of the model on long-tail data and undersampled data. On large-scale data, many (or most) tasks and domains may be represented by few (important) datapoints. During normal training, those datapoints are highly influential for the behavior of the model on their subpopulation. However, this is not acceptable for DP training, leading to these datapoints being effectively discounted. This may reduce accuracy of the model on long-tail tasks (a main selling point of Foundational Models) and on under-represented populations (a fairness risk). In fact, some have advanced the notion that \textit{memorization} of the long-tails (the opposite of DP) is essential for good performance of ML models \cite{feldman2020does}, in some cases provably so \cite{brown2021memorization}.

Ultimately, Model Disgorgement is only a small portion of a comprehensive process of ensuring that AI Models are trained with data that satisfies all legal constraints and that gives content owners and creators proper attribution for their data. Careful data collection, curation, documentation, provenance, and licensing are always the first starting point. Model Disgorgement is not intended as a substitute for those, but rather an additional option if the need or desire to eliminate a cohort of data and their effects becomes manifest. Some of the methods we describe, referred to as preemptive, also provide means for content owners and creators to understand and leverage the way their data is used in order to properly recognize the value of their data upon licensing.

\bibliographystyle{plain}
\bibliography{bibliography}

\begin{thebibliography}{10}

\bibitem{achille2019information}
Alessandro Achille, Giovanni Paolini, and Stefano Soatto.
\newblock Where is the information in a deep neural network?
\newblock {\em arXiv preprint arXiv:1905.12213}, 2019.

\bibitem{achille2022binding}
Alessandro Achille and Stefano Soatto.
\newblock On binding objects to symbols: Learning physical concepts to
  understand real from fake.
\newblock {\em arXiv preprint arXiv:2207.12186}, 2022.

\bibitem{bansal2019updates}
Gagan Bansal, Besmira Nushi, Ece Kamar, Daniel~S Weld, Walter~S Lasecki, and
  Eric Horvitz.
\newblock Updates in human-ai teams: Understanding and addressing the
  performance/compatibility tradeoff.
\newblock In {\em Proceedings of the AAAI Conference on Artificial
  Intelligence}, volume~33, pages 2429--2437, 2019.

\bibitem{basu2020influence}
Samyadeep Basu, Philip Pope, and Soheil Feizi.
\newblock Influence functions in deep learning are fragile.
\newblock {\em arXiv preprint arXiv:2006.14651}, 2020.

\bibitem{sisa}
Lucas Bourtoule, Varun Chandrasekaran, Christopher~A. Choquette-Choo, Hengrui
  Jia, Adelin Travers, Baiwu Zhang, David Lie, and Nicolas Papernot.
\newblock Machine unlearning.
\newblock In {\em 2021 IEEE Symposium on Security and Privacy (SP)}, pages
  141--159, 2021.

\bibitem{brown2021memorization}
Gavin Brown, Mark Bun, Vitaly Feldman, Adam Smith, and Kunal Talwar.
\newblock When is memorization of irrelevant training data necessary for
  high-accuracy learning?
\newblock In {\em Proceedings of the 53rd annual ACM SIGACT symposium on theory
  of computing}, pages 123--132, 2021.

\bibitem{cao2015towards}
Yinzhi Cao and Junfeng Yang.
\newblock Towards making systems forget with machine unlearning.
\newblock In {\em 2015 IEEE Symposium on Security and Privacy}, pages 463--480,
  2015.

\bibitem{dong2022gaussian}
Jinshuo Dong, Aaron Roth, and Weijie~J Su.
\newblock Gaussian differential privacy.
\newblock {\em Journal of the Royal Statistical Society Series B: Statistical
  Methodology}, 84(1):3--37, 2022.

\bibitem{inca}
Yonatan Dukler, Alessandro Achille, Hao Yang, Varsha Vivek, Luca Zancato, Ben
  Bowman, Avinash Ravichandran, Charless Fowlkes, Ashwin Swaminathan, and
  Stefano Soatto.
\newblock Introspective cross-attention probing for lightweight transfer of
  pre-trained models.
\newblock {\em arXiv preprint arXiv:2303.04105}, 2023.

\bibitem{bowman2023safe}
Yonatan Dukler, Benjamin Bowman, Alessandro Achille, Aditya Golatkar, Ashwin
  Swaminathan, and Stefano Soatto.
\newblock Safe: Machine unlearning with shard graphs.
\newblock preprint, 2023.

\bibitem{feldman2020does}
Vitaly Feldman.
\newblock Does learning require memorization? a short tale about a long tail.
\newblock In {\em Proceedings of the 52nd Annual ACM SIGACT Symposium on Theory
  of Computing}, pages 954--959, 2020.

\bibitem{ginart2019making}
Antonio Ginart, Melody Guan, Gregory Valiant, and James~Y Zou.
\newblock Making ai forget you: Data deletion in machine learning.
\newblock {\em Advances in neural information processing systems}, 32, 2019.

\bibitem{mixedprivacyforgettinggolatkar}
Aditya Golatkar, Alessandro Achille, Avinash Ravichandran, Marzia Polito, and
  Stefano Soatto.
\newblock Mixed-privacy forgetting in deep networks.
\newblock In {\em Proceedings of the IEEE/CVF Conference on Computer Vision and
  Pattern Recognition (CVPR)}, pages 792--801, June 2021.

\bibitem{Golatkar_2020_CVPR}
Aditya Golatkar, Alessandro Achille, and Stefano Soatto.
\newblock Eternal sunshine of the spotless net: Selective forgetting in deep
  networks.
\newblock In {\em Proceedings of the IEEE/CVF Conference on Computer Vision and
  Pattern Recognition (CVPR)}, June 2020.

\bibitem{golatkar2020forgetting}
Aditya Golatkar, Alessandro Achille, and Stefano Soatto.
\newblock Forgetting outside the box: Scrubbing deep networks of information
  accessible from input-output observations.
\newblock In {\em European Conference on Computer Vision}, pages 383--398.
  Springer, 2020.

\bibitem{golatkar2022mixed}
Aditya Golatkar, Alessandro Achille, Yu-Xiang Wang, Aaron Roth, Michael Kearns,
  and Stefano Soatto.
\newblock Mixed differential privacy in computer vision.
\newblock In {\em Proceedings of the IEEE/CVF Conference on Computer Vision and
  Pattern Recognition}, pages 8376--8386, 2022.

\bibitem{harutyunyan2021estimating}
Hrayr Harutyunyan, Alessandro Achille, Giovanni Paolini, Orchid Majumder,
  Avinash Ravichandran, Rahul Bhotika, and Stefano Soatto.
\newblock Estimating informativeness of samples with smooth unique information.
\newblock {\em arXiv preprint arXiv:2101.06640}, 2021.

\bibitem{heusel2017gans}
Martin Heusel, Hubert Ramsauer, Thomas Unterthiner, Bernhard Nessler, and Sepp
  Hochreiter.
\newblock Gans trained by a two time-scale update rule converge to a local nash
  equilibrium.
\newblock {\em Advances in neural information processing systems}, 30, 2017.

\bibitem{kochno}
Korbinian Koch and Marcus Soll.
\newblock No matter how you slice it: Machine unlearning with sisa comes at the
  expense of minority classes.
\newblock In {\em First IEEE Conference on Secure and Trustworthy Machine
  Learning}.

\bibitem{koh2017understanding}
Pang~Wei Koh and Percy Liang.
\newblock Understanding black-box predictions via influence functions.
\newblock In {\em International conference on machine learning}, pages
  1885--1894. PMLR, 2017.

\bibitem{kumar2022privacy}
Vinayshekhar~Bannihatti Kumar, Rashmi Gangadharaiah, and Dan Roth.
\newblock Privacy adhering machine un-learning in nlp.
\newblock {\em arXiv preprint arXiv:2212.09573}, 2022.

\bibitem{moosavi2017universal}
Seyed-Mohsen Moosavi-Dezfooli, Alhussein Fawzi, Omar Fawzi, and Pascal
  Frossard.
\newblock Universal adversarial perturbations.
\newblock In {\em Proceedings of the IEEE conference on computer vision and
  pattern recognition}, pages 1765--1773, 2017.

\bibitem{radford2021learning}
Alec Radford, Jong~Wook Kim, Chris Hallacy, Aditya Ramesh, Gabriel Goh,
  Sandhini Agarwal, Girish Sastry, Amanda Askell, Pamela Mishkin, Jack Clark,
  et~al.
\newblock Learning transferable visual models from natural language
  supervision.
\newblock In {\em International conference on machine learning}, pages
  8748--8763. PMLR, 2021.

\bibitem{rensink2001change}
Ronald~A Rensink.
\newblock Change blindness: Implications for the nature of visual attention.
\newblock {\em Vision and attention}, pages 169--188, 2001.

\bibitem{shan2023glaze}
Shawn Shan, Jenna Cryan, Emily Wenger, Haitao Zheng, Rana Hanocka, and Ben~Y
  Zhao.
\newblock Glaze: Protecting artists from style mimicry by text-to-image models.
\newblock {\em arXiv preprint arXiv:2302.04222}, 2023.

\bibitem{shen2020towards}
Yantao Shen, Yuanjun Xiong, Wei Xia, and Stefano Soatto.
\newblock Towards backward-compatible representation learning.
\newblock In {\em Proceedings of the IEEE/CVF Conference on Computer Vision and
  Pattern Recognition}, pages 6368--6377, 2020.

\bibitem{srivastava2020empirical}
Megha Srivastava, Besmira Nushi, Ece Kamar, Shital Shah, and Eric Horvitz.
\newblock An empirical analysis of backward compatibility in machine learning
  systems.
\newblock In {\em Proceedings of the 26th ACM SIGKDD International Conference
  on Knowledge Discovery \& Data Mining}, pages 3272--3280, 2020.

\bibitem{wang2004image}
Zhou Wang, Alan~C Bovik, Hamid~R Sheikh, and Eero~P Simoncelli.
\newblock Image quality assessment: from error visibility to structural
  similarity.
\newblock {\em IEEE transactions on image processing}, 13(4):600--612, 2004.

\bibitem{wu2020unsupervised}
Yuhao Wu, Weiping Ji, and Jinjian Wu.
\newblock Unsupervised deep learning for just noticeable difference estimation.
\newblock In {\em 2020 IEEE International Conference on Multimedia \& Expo
  Workshops (ICMEW)}, pages 1--6. IEEE, 2020.

\bibitem{xie2020unsupervised}
Qizhe Xie, Zihang Dai, Eduard Hovy, Thang Luong, and Quoc Le.
\newblock Unsupervised data augmentation for consistency training.
\newblock {\em Advances in neural information processing systems},
  33:6256--6268, 2020.

\bibitem{arcane}
Haonan Yan, Xiaoguang Li, Ziyao Guo, Hui Li, Fenghua Li, and Xiaodong Lin.
\newblock Arcane: An efficient architecture for exact machine unlearning.
\newblock In Lud~De Raedt, editor, {\em Proceedings of the Thirty-First
  International Joint Conference on Artificial Intelligence, {IJCAI-22}}, pages
  4006--4013. International Joint Conferences on Artificial Intelligence
  Organization, 7 2022.
\newblock Main Track.

\bibitem{yan2021positive}
Sijie Yan, Yuanjun Xiong, Kaustav Kundu, Shuo Yang, Siqi Deng, Meng Wang, Wei
  Xia, and Stefano Soatto.
\newblock Positive-congruent training: Towards regression-free model updates.
\newblock In {\em Proceedings of the IEEE/CVF Conference on Computer Vision and
  Pattern Recognition}, pages 14299--14308, 2021.

\bibitem{zeng2023towards}
Yi~Zeng, Zhouxing Shi, Ming Jin, Feiyang Kang, Lingjuan Lyu, Cho-Jui Hsieh, and
  Ruoxi Jia.
\newblock Towards robustness certification against universal perturbations.
\newblock In {\em The Eleventh International Conference on Learning
  Representations}, 2023.

\bibitem{zhao2020sim}
Wenshuai Zhao, Jorge~Pe{\~n}a Queralta, and Tomi Westerlund.
\newblock Sim-to-real transfer in deep reinforcement learning for robotics: a
  survey.
\newblock In {\em 2020 IEEE symposium series on computational intelligence
  (SSCI)}, pages 737--744. IEEE, 2020.

\bibitem{legonet}
Xiaofei Zhu, Jie Wu, Ling Zhu, Jiafeng Guo, Ran Yu, Katarina Boland, and Stefan
  Dietze.
\newblock Exploring user historical semantic and sentiment preference for
  microblog sentiment classification.
\newblock {\em Neurocomputing}, 464:141--150, 2021.

\end{thebibliography}

\end{document}